\documentclass[letterpaper, 10 pt, conference]{ieeeconf}  

\IEEEoverridecommandlockouts                              

\usepackage{graphicx}
\usepackage{caption}

\usepackage{comment}
\usepackage{booktabs}
\usepackage{amsmath} 
\usepackage{amsfonts}
\usepackage{subfig}
\usepackage{bm}
\usepackage{algorithm}
\usepackage[noend]{algpseudocode}
\let\labelindent\relax
\usepackage[]{enumitem}

\usepackage[font=small,labelfont=bf]{caption}
\usepackage{cite}

\DeclareMathOperator*{\argmax}{arg\,max}

\newcommand{\R}{\mathbb{R}}

\newcommand{\etc}{\textit{etc}. }

\title{\LARGE \bf
Collaborative Planning for Mixed-Autonomy Lane Merging
}

\author{Shray Bansal$^{1}$, Akansel Cosgun$^{2}$, Alireza Nakhaei$^{3}$ and Kikuo Fujimura$^{3}$
 \thanks{$^{1}$S. Bansal is with Georgia Institute of Technology, Atlanta, GA, USA
 {\tt\small sbansal34@gatech.edu}}%
 \thanks{$^{2}$A. Cosgun is with Monash University, Clayton, VIC, Australia}%
 \thanks{$^{3}$A. Nakhaei and K. Fujimura are with Honda Research Institute, Mountain View, CA, USA}%
}

\begin{document}

\setlength{\abovedisplayskip}{3pt}
\setlength{\belowdisplayskip}{3pt}

\maketitle
\thispagestyle{empty}
\pagestyle{empty}

\begin{abstract}

Driving is a social activity: drivers often indicate their intent to change lanes via motion cues. We consider mixed-autonomy traffic where a Human-driven Vehicle (HV) and an Autonomous Vehicle (AV) drive together. We propose a planning framework where the degree to which the AV considers the other agent's reward is controlled by a selfishness factor. We test our approach on a simulated two-lane highway where the AV and HV merge into each other's lanes. In a user study with 21 subjects and 6 different selfishness factors, we found that our planning approach was sound and that both agents had less merging times when a factor that balances the rewards for the two agents was chosen. Our results on double lane merging suggest it to be a non-zero-sum game and encourage further investigation on collaborative decision making algorithms for mixed-autonomy traffic.

\end{abstract}

\section{Introduction}
\label{sec:introduction}

Driving is a social activity: drivers indicate their willingness to change lanes by subtle cues such as eye contact, or by not-so-subtle cues such as adjusting their speed and position \cite{ba2015effect}. There has been impressive demonstrations 

of Autonomous Vehicle (AV) technology \cite{cosgun2017towards, urmson2008autonomous, leonard2008perception}, however one of the remaining challenges in this area is reading those cues to estimate the intentions of other agents as well as using cues to communicate the intentions of the AV. As AVs become commonplace, the situations where AV's and Human-driven Vehicles (HV) interact will increase. A number of issues in mixed-autonomy traffic need to be addressed before wide deployment, many posing interesting  technical challenges.

\begin{figure}[ht!]
\centering
\includegraphics[width=0.3\linewidth]{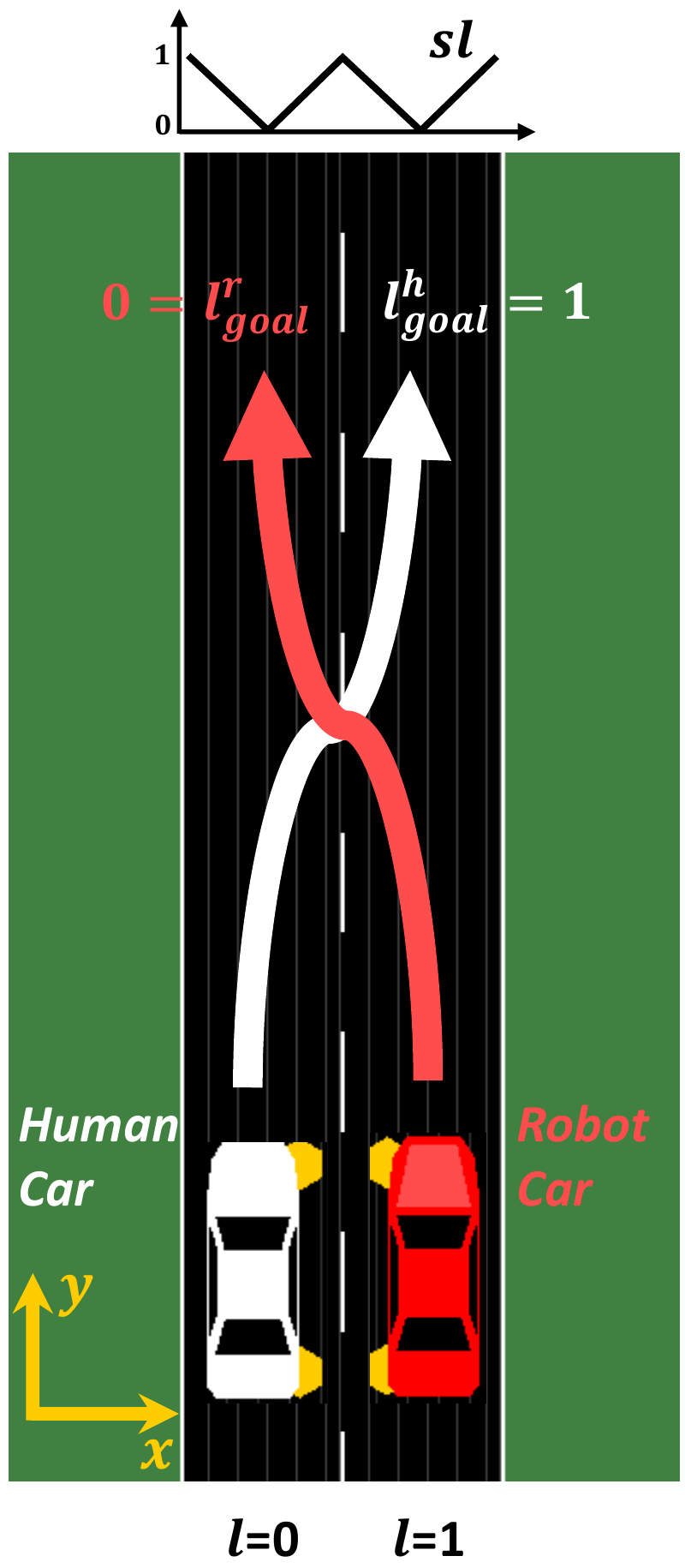}
\caption{Double lane merging problem. The autonomous and human-driven vehicles start together and aim to switch lanes. Successful merging requires signaling the agent's own intent while estimating the other's intention and then acting on it.}\label{fig:problem_setup}
\vspace{-0.2cm}
\end{figure}

Prior work focused on designing robust controllers for low-level tasks such as lane following and lane changing
\cite{urmson2008autonomous,leonard2008perception, petrov2013adaptive, hatipoglu2003automated,shiller1998emergency} either considered the other drivers as obstacles to avoid \cite{shiller1998emergency,petrov2013adaptive} or did not consider their presence while modeling \cite{hatipoglu2003automated}.
Close interactions with other agents in high-level tasks like deciding when to pass or change lanes 
\cite{matthews2001model,sukthankar1998multiple} require a more sophisticated model. 
For instance, a highway merge where there is a short distance to the next exit creates a situation where cars entering and exiting have to negotiate with one another to merge safely into their desired lanes. In this paper, we study a planning approach for navigating an autonomous vehicle for this challenging double lane merge in the presence of a human-driven car, see Fig. \ref{fig:problem_setup}. 

Recent progress in Reinforcement Learning (RL) has led to work exploring its application to autonomous driving \cite{isele2018navigating,shalev2016safe}. However, most of this progress was for the single-agent setting but driving is inherently multi-agent. \cite{isele2018navigating} solve the issue by only considering the other agent's actions through statistics of the traffic and although, \cite{shalev2016safe} study the double lane merging problem in a multi-agent framework, they simplify the problem using expert knowledge. 
Similar to our approach of online planning, RL with Monte Carlo tree search \cite{silver2016mastering} has been applied to autonomous driving in \cite{bouton2017belief}.

The idea of an autonomous agent acting deliberately to be predictable was formalized by \cite{dragan2013legibility}. Game theoretic models have been used to model human's adaptation to the robot \cite{nikolaidis2017game} in the human-robot interaction domain. We are also influenced by research on collaborative planning with pedestrians for navigating a mobile robot in a social environment 
\cite{trautman2010unfreezing, cosgun2016anticipatory}. Our work is most related to \cite{sadigh2016planning}, which leverages the adaptability of the human by being cognizant of the effect the AV's actions have on the human driver. They show that the AV can influence a human driver such as making them slow down by moving in front of them. 

However, while their approach maximizes the AV's own reward, it is not uncommon to see people slow down and let others into their lane while driving.
Inspired by this idea of consideration and of leveraging the adaptability of HVs, we frame mixed-autonomy driving as a collaboration. In particular, we consider the double lane merging problem where the AV and HV start in adjacent lanes and must merge into each others' lanes in a limited road length. The actions of the AV are considered explicitly via collaborative planning with those of the HV to optimize a collective reward that combines the rewards of both of the agents. We validate our planning approach in simulation and conduct a user study where subjects can engage in interactions with our autonomous agents. We study the effect of the AV having different levels of selfishness with users. 

We make the following two contributions: 
\begin{itemize}
\item A mixed-autonomy merge planning algorithm that is able to vary its consideration for other drivers with a single tunable selfishness factor.
\item Demonstration that an AV with balanced consideration leads to better performance for both agents, suggesting that lane merging is a non-zero sum game.
\end{itemize}

\section{PROBLEM}
\label{sec:problem}

This paper assumes a discrete-time Markov Decision Process defined by,
\begin{equation}
M = \langle S, U, T, R\rangle,
\end{equation}
where $S \subseteq \R^n$ is a $n$-dimensional state space, 
$U = \{1,...,K\}$
is the set of discrete actions the system can perform, $T: S\times U \to S$ is a deterministic transition function, and $R: S \to \R$ is a reward function.
\subsection{State Representation}
The state representation $s$ provides full information of the vehicles present in the system. It contains the continuous lateral $x$ and longitudinal positions $y$, as well as the longitudinal speed $v$ of all cars including the AV.
\begin{equation}
s = (y_i, x_i, v_i)_{i \in \{ 1, .. , k\}}
\label{eq:state}
\end{equation}
Where $k$ is the number of cars in the environment. Since the focus of this study is on decision-making pertaining to intelligent interaction we limit the scope to the fully-observable setting. 

\subsection{Action Representation}
The action representation $u$ contains actions for each car $u = (u_i)_{i \in \{ 1, .. , k\}}$. Each car has five control actions available to them at every time-step:
\begin{itemize}
\item \textbf{accelerate}: increases the speed by a constant factor. 
\item \textbf{decelerate}: decreases the speed by a constant factor. 
\item \textbf{stay}: maintains the speed.
\item \textbf{turn-right}: moves the car in the positive latitudinal direction while reducing its longitudinal motion.
\item \textbf{turn-left}: moves the car in the negative latitudinal direction while reducing its longitudinal motion.
\end{itemize}

The \textbf{right} and \textbf{left} actions are not permitted if the car is at the rightmost and leftmost edges of the road respectively. These discrete high-level actions were chosen because the article aims to study the high-level decision-making of the cars and so a controller is assumed.

\subsection{Dynamics}
\label{ssec:dynamics}
The state at the next time-step is determined by applying a control action to each car in the system $s_{t+1} = T(s_t, u)$.
These actions enable longitudinal and lateral control of the car and are applied instantaneously. The dynamics follow a simple deterministic model for each car. The accelerate, decelerate and stay actions change the state of the corresponding vehicle in the following way. 
\begin{align}
        v_{t+1} &= v_t + a \Delta t \\[1pt]
        y_{t+1} &= y_t + v_t \Delta t \\[1pt]
        x_{t+1} &= x_t
\end{align}

Here, $\Delta t$ is the length of the time-step and $a$ is the acceleration. For the turning actions we introduce a fixed lateral velocity ($v_{lat}$), 
\begin{align}
v^x_t &= min \{v_t, v_{lat}\} \\[1pt]
v^y_t &= \sqrt{v_t^2 - {v^x_t}^2} \\[1pt]
y_{t+1} &= y_t + v^y_t \Delta t \\[1pt]
x_{t+1} &= x_t + v^x_t \Delta t \\[1pt]
v_{t+1} &= v_t
\end{align}

\subsection{Problem Domain}
We use the two-lane scenario depicted in Fig. \ref{fig:problem_setup} where the HV and AV start in adjacent lanes and have the goal of merging into each other's lanes before the road ends. Here, if the cars change lanes independent (without regard) of each other it can cause a collision, which leads to the interesting challenge of planning while cognizant of the human driver's goals. The investigation is restricted to a 
two car scenario, with one human driver ($H$) and the other an AV, which will be referred to as the robot ($R$) car. We do so to reduce the effect of other interactions like HV-HV and AV-AV, and make the effects of the focus of our study - interactions between HVs and AVs - more clearly observable.

Even though our approach is extensible to multiple cars, we limit the system to two cars in this discussion and will apply a subscript of $R$ or $H$ to the AV and human, respectively. Thus, the state $s$ is composed of the state for the human $s_H$ and that of the robot $s_R$, where $s_H = (y_H, x_H, v_H)$ and $s_R= (y_R, x_R, v_R)$.
We assume that the AV has noise-free access to the state at the current time $s^t$ and the human's goal.

\subsection{Reward}
The goal of the AV is to reach its target lane ($l_{goal}$) while avoiding collisions and preferring to drive in the center of the lane. We encoded this high-level goal into a simple reward function that gets applied at every time-step. It has a high negative reward for collisions and a positive one for being in the goal-lane. To keep the car near the center of the lane, we add a small reward that decays exponentially with distance to the center.   
\begin{equation}
\label{eq:reward}
r(s) = \begin{cases}
-10, & \text{if collision} \\
\gamma e^{-sl} + (1-\gamma), & \text{if } l = l_{goal} \\
0, & \text{if } l \neq l_{goal}
\end{cases}
\end{equation}

Here, $l = \{0,1\}$ and $sl = [0, 1]$ are the lane and sublane positions of the robot, explained in Fig. \ref{fig:problem_setup}, and, $\gamma = 0.3$ is a constant used to make the agent favor the middle of the lane after it reaches the target lane. The positions $l$ and $sl$ can be easily determined from the latitudinal position $x$. 

\section{APPROACH}
\label{sec:approach}

The goal of the AV is to perform a set of actions that optimize the reward it receives over time. To do so, it will plan for a finite sequence of actions, perform the first one and re-plan after receiving an observation.
Let $N$ be the length of the sequence, $\bm{u_R} = (u_R^t)_{t=0}^{N-1}$ 

be a sequence of the AV's actions and $\bm{u_H} = (u_H^t)_{t=0}^{N-1}$ be those of the human. Then the accumulated reward for the AV over the sequence starting at a state $s^t$ is given by
\begin{equation}
	R_R(s^t, \bm{u_R}, \bm{u_H}) = \sum_{i=0}^{N-1}{r_R(T(s^{t+i}, u_H^{t+i}, u_R^{t+i}))}.
\end{equation}
Here, $r_R$ is the instantaneous reward defined in (\ref{eq:reward}) and the next state $s^{t+1} = T(s^{t}, u_H^{t}, u_R^{t})$ is computed using the dynamics $T$ described in Section \ref{ssec:dynamics}. 
From a state $s^t$, the optimal set of actions for the AV, $\bm{u_R^*}$, can be found through the following optimization.   
\begin{equation}
\label{egn:robot_opt}
\bm{u_{R}^*} = \argmax_{\bm{u_R}} R_R(s^t, \bm{u_R}, \bm{u_H})
\end{equation}
In the ideal scenario, $\bm{u_H}$ would hold the HV's actual future actions, however, these are unknown. Some previous work presumed that $\bm{u_H}$ depends only on the state, for instance, \cite{vitus2013probabilistic} uses a constant velocity model, but a more accurate model for a rational driver would also be influenced by the AV's future plan. We model this influence by assuming that the human driver plans in an action space that includes the AV as well and chooses to optimize a reward which includes the goals of both. This is based on the hypothesis that driving is not a zero-sum game and people drive with some consideration of each other's goals. In this system, we are indirectly influencing the human agent's actions by choosing the reward function for this joint planning system. 

\subsection{Planning and Prediction for Collaboration}
Our collaborative planning approach determines the optimal set of actions by maximizing the joint reward $R_J$ which is a weighted combination of the human and robot rewards, 
\begin{equation}
R_J = \alpha R_R + (1-\alpha) R_H.
\end{equation}

Here, $\alpha$ is the selfishness factor which determines the relative importance of each reward in the collaboration. For instance, $\alpha=1$ is an AV with no consideration for the human and $\alpha=0$ is the opposite, while $\alpha=0.5$ equally considers the goal of both agents. We do not assume apriori knowledge of $\alpha$ and plan to study its effect with the user study. The optimal plan for both agents under this model is determined by the following optimization.
\begin{equation}
\label{egn:collab_opt}
\bm{u_{R}^*}, \bm{u_{H}^*} = \argmax_{\bm{u_R}, \bm{u_H}} R_J(s, \bm{u_R}, \bm{u_H})
\end{equation}

Since we do not control the human's car, the set of actions $\bm{u_{H}^*}$ is a prediction of their plan under the influence of the AV's plan $\bm{u_R^*}$. This influence is implicit in our model through choice of $R_J$.
The AV is controlled by executing the first action from $\bm{u_{R}^*}$ and the planning is repeated after every time-step. This planning is performed by an online search procedure described next. 

\subsection{Limited Horizon Tree Search}

The optimization from (\ref{egn:collab_opt}) is implemented as an online tree-search. Here, the nodes are states containing both human and AV positions and velocities, from (\ref{eq:state}). The search starts at the current state and then simulates taking actions, terminating when either all the unpruned nodes, until maximum depth, are explored or if the time limit is reached. The time limit is set to be the length of the simulation time-step for real-time planning. The action leading to the maximally rewarding node is chosen and ties are broken randomly.

The algorithm is similar to $A^*$ search \cite{russell2003artificial} while also handling multiple unknown goal nodes and a simplified version is provided in $\textsc{find\_optimal\_action()}$ of Algorithm \ref{algo:search}. The current state $s^{init}$ and maximum depth $t_{max}$ are used as input, a priority queue $open$ keeps track of the states to explore and a set $closed$ of those that have already been explored. States are ordered by $f$, which is the sum of their accumulated reward $R$ and the heuristic. $\textsc{get\_reward()}$ returns the reward at a state (\ref{eq:reward}), while $rewards[s]$ maps an explored state to its accumulated reward. $\textsc{is\_terminal()}$ checks if a given state is a terminal one, and $\textsc{get\_heuristic()}$ provides an upper bound by returning the maximum reward in the remaining horizon (or depth), keeping the search optimal. The $\textsc{timer}()$ tracks the amount of time spent in the search and outputs true in case the time budget runs out. We also keep track of parent states and actions (not shown) to efficiently trace-back optimal actions. In case search is terminated prematurely, the state with maximal accumulated reward is chosen. 

\begin{algorithm}
    \begin{algorithmic}[1]
    	\Procedure{find\_optimal\_action}{$s^{init}, t_{max}$}
        \State Initialize $open, closed, rewards$ as $\emptyset$
        \State $t \gets 0$
        \State Push $((s^{init}, 0), \Call{get\_heuristic}{t_{max}-0})$ into $open$
        \State $R_{max} \gets -\infty,  $
        \While{$open$ not empty}
        	\State \textbf{break if} $\Call{timer}$
        	\State $(s, t), f \gets \Call{pop}{open}$ 
            \State \textbf{break} if $f < R_{max}$ \Comment{Pruning} \label{algo:search:prune}
            \State \textbf{add} $s$ to $closed$
            \If{\Call{is\_terminal}{s, t}}
            	\If{$rewards[s] > R_{max}$}
                	\State $R_{max} \gets rewards[s]$ 
                \EndIf
         		\State \textbf{continue}
            \EndIf
            \State $t \gets t+1$
        	\ForAll{$s_+ \in \Call{get\_neighbors}{s}$}
            	\State $R \gets rewards[s] + \Call{get\_reward}{s_+}$
                \If {($s_+ \in closed) \land (R < rewards[s_+]$)}
                	\State \textbf{continue}
                \EndIf
                \State $f_+ \gets R + \Call{get\_heuristic}{t_{max}-(t+1)}$
                \State $rewards[s_+] \gets R$
                \State Push $((s_+,t+1), f_+)$ into $open$
        	\EndFor
        \EndWhile
        \State Traceback from state with reward $R_{max}$ to find optimal action $(u^*_H, u^*_R)$
        \State \textbf{return} $u^*_R$
        \EndProcedure
        \Procedure{get\_heuristic}{$t_{remaining}$}
        	\State \textbf{return} $t_{remaining} \cdot r_{max}$
        \EndProcedure
    \end{algorithmic}
    \caption{Finite horizon action search}
    \label{algo:search}
\end{algorithm}

\subsection{Selfish Baseline}
\label{sub:baseline}
Inspiring work from \cite{sadigh2016planning} had a similar formulation for mixed-autonomy driving. However, they determine the AV's plan $\bm{u_R^*}$ from (\ref{egn:robot_opt}) by optimizing $R_R$ only and the human's plan, $\bm{u_H}$, is determined by optimizing $R_H$, leading to the following nested optimization. 
\begin{equation}
\label{eq:sadigh}
\bm{u_{R}^*} = \argmax_{\bm{u_R}} R_R(s, \bm{u_R}, (
\argmax_{\bm{u_H}} R_H(s, \bm{\tilde{u}_R}, \bm{u_H})),
\end{equation}

To solve it they make some simplifying assumptions. They assume a turn-taking scenario where the robot plans first, also they give the human model access to $\bm{u_R}$, i.e. $\bm{\tilde{u}_R} = \bm{u_R}$.
This leads to the human considering the robot's plan as fixed when planning her response. 
The reward $R_R$ in our task depends only upon the state of the AV apart from avoiding collisions, reducing (\ref{eq:sadigh}) to the selfish $\alpha=1$ condition of our approach because the human-model here considers the plan $\bm{u_R}$ as fixed. So, the only case that can cause a $\bm{u_R}$ that optimizes $R_J=R_R$ in (\ref{egn:collab_opt}) to be suboptimal in (\ref{eq:sadigh}) is when the optimal $u_H$ causes a collision. This, however, is prevented by the high negative reward of collisions for $R_H$. 


This formulation works well in scenarios where $R_R$ is designed to encode desired human behavior explicitly. For instance, they included the negative squared velocity of the human into $R_R$ to reward the AV for slowing down the human. However, reward functions encoding the agent's own goals, rather than those of other drivers, are more intuitive, and there, this formulation might lead to selfish behaviors as shown in our experiments.

\subsection{Implementation Details}
The human reward $R_H$ was chosen to have the same simple form as the robot (\ref{eq:reward}). Our results indicate that the model performs well even with this approximation of the human's reward. However, it can easily be replaced by a learned model in our framework in case sequences for learning this reward are available, like in \cite{sadigh2016planning}. The search algorithm \ref{algo:search}, was implemented in Python. For the simulation we used the open source Simulation of Urban MObility (SUMO) \cite{krajzewicz2012recent} software and its in-built visualizer to render the cars. We used a discrete-time simulation with a time-step of $0.2$ seconds, which also served as the time-limit for the search algorithm to return an action. The cars were allowed a maximum longitudinal speed of $30 m/s$ and a lateral speed of $3 m/s$, with a lane width of $4m$. 
If the cars start together with similar speeds it can take multiple seconds even for the optimal plan, to separate the cars enough so that a lane merge can commence safely. Thus, the time horizon of the search was kept at $6$ seconds for interesting plans to develop. However, with a $0.2s$ time-step this would mean searching a tree of depth $30$ and branching factor of $5^2$ this leads to exploring $25^{30}$ states which is infeasible in $0.2s$. To make this search faster we increased the time-step to $1s$ in the planner, which is equivalent to repeating each action for five time-steps. This, combined with our choice of data-structures enabled us to typically complete the search in less than $0.2s$. 

\vspace{-0.2cm}
\section{USER STUDY}
\label{sec:userstudy}

A human-subject study was conducted to evaluate our method on the double lane merge scenario described earlier.  

\begin{figure}
\centering
\includegraphics[width=0.9\linewidth]{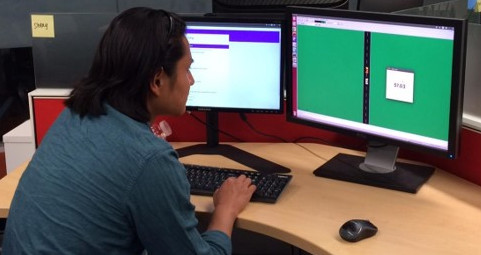}
\caption{A participant in our user study drives the car using the keyboard. }
\label{fig:user_study}
\vspace{-0.15cm}
\end{figure}

\subsection{Independent Variables}
We manipulated two aspects of this experiment:
\begin{enumerate}
\item \textbf{Road Length.} We used two lengths of the road $\{100, 200\}$m referred to as \textit{short-road} and \textit{long-road}.
This creates two levels of difficulty for the experiment where the more challenging \textit{short road} scenario gives the HV and AV less time to complete their merges.
\item \textbf{Selfishness Factor.} The selfishness factor ($\alpha$), defined earlier, was given six different values, \[\alpha \in \{0.0, 0.2, 0.4, 0.6, 0.8, 1.0\}\] which will be referred to as \textit{0R, .2R, .4R, .6R, .8R, 1R} respectively, where the level of the AV's selfishness increases with increasing $\alpha$. 
We consider the \textit{.4R}, \textit{.6R} agents as fair and \textit{1R} agent as selfish in the discussion. 
\end{enumerate}
Through these we explored a total of $2 \times 6 = 12$ conditions.

\subsection{Hypotheses}
The following hypotheses are tested.
\begin{enumerate}[label=\textbf{H\arabic*}]

\item \textit{The selfishness factor will have an impact on the human's performance on the task.} In particular, we expect the less selfish versions of our algorithm to help improve the human driver's performance.


\item \textit{Decreasing selfishness to a fair balance of rewards will not adversely effect the AV's performance.} We take \textit{.4R} and \textit{.6R} to be fair versions of our approach and expect that their performance is not significantly worse than the selfish \textit{1R} agent.

\end{enumerate}

\subsection{Experimental Design}

We recruited $21$ participants, out of which $4$ were female and $17$ male, $19$ had a valid driver's license (and were kept), and $20$ were between the ages of $21$ and $40$ with one older. 

The subjects had a top-down view of the environment and controlled the car using a keyboard, see Fig. \ref{fig:user_study}. Each participant was instructed to drive their car to the goal lane, in a safe and natural manner, before the end of the road. A small window displayed the distance to the end of the road. There was one other agent (AV) present in the scene that used indicator-lights to communicate its goal lane. We did not inform the participants that this agent was autonomously controlled to avoid biasing the perception of it.
An unrecorded practice scenario was used to familiarize them with the simulation environment and the controls of their car. A within-subject design was used to mitigate the effects of inter-subject variability. Each subject performed $18$ trials, each of which was uniformly sampled from one of the $12$ conditions mentioned before. After every trial, a questionnaire gauged their interaction with the robot on a three-point scale. The sequences and responses of these trials were recorded and analyzed. 
To avoid making the task repetitive and losing the user's interest, we varied start lanes and the color of the robot's car through uniform sampling. The robot's initial speed was sampled from a normal distribution with a mean $15 m/s$ and $3 m/s$ standard deviation, while the human car's was kept constant at $15 m/s$. The subjects were not informed of the variables being manipulated. 

\section{RESULTS AND ANALYSIS}
\label{sec:results}
\subsection{Objective Measures}

\begin{figure}[t!]
\begin{tabular}{cc}
\hbox{\hspace{-1.2em} \includegraphics[width=0.5\linewidth]{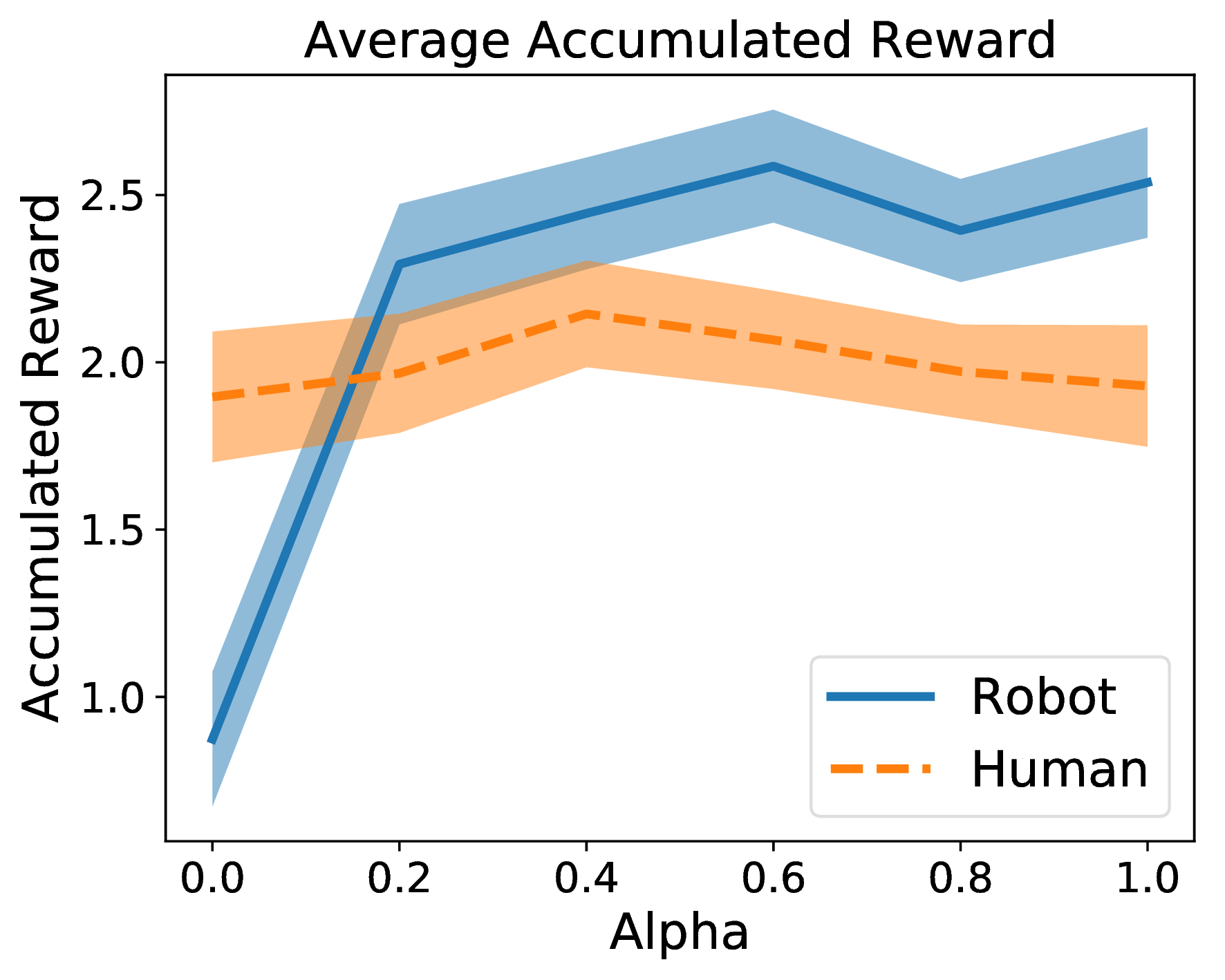}}
&   
\hbox{\hspace{-1.2em} \includegraphics[width=0.5\linewidth]{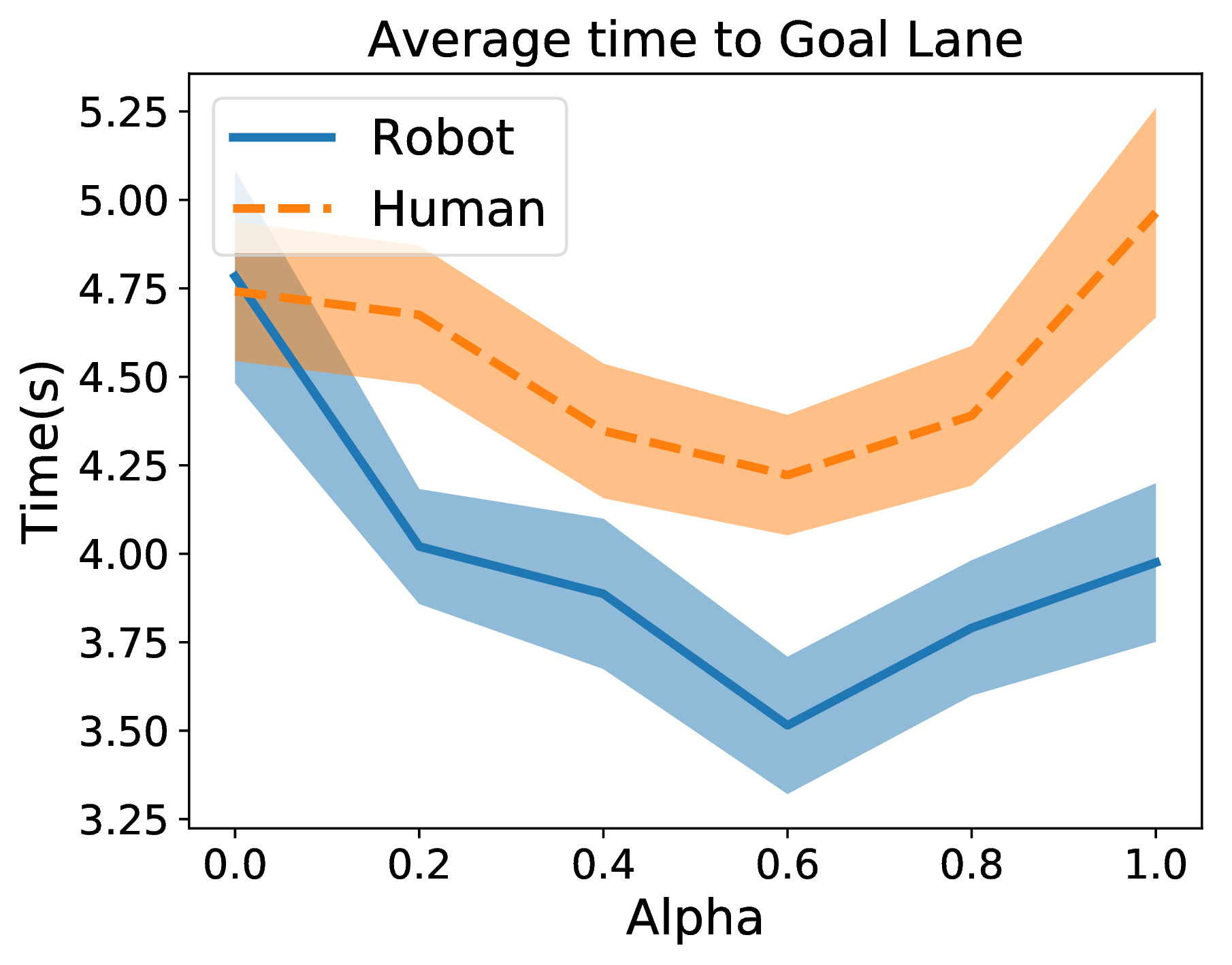}}
\\
(a)  & (b)
\end{tabular}
\caption{Quantitative effect of $\alpha$. (a) Plots the average reward accumulated by both agents and (b) shows the time it takes for a car to merge into its goal lane under all six conditions of the cooperation factor $\alpha$. Fig. shows the mean as the line and the standard errors as the region. We observe that the fair conditions \emph{.4R, .6R} fare better than others including the selfish \emph{1R} for both agents. Note that higher rewards and lower merge-times relate to better performance.
}
\label{fig:timing_reward}
\end{figure}

We studied the effect of $\alpha$, using our $325$ recorded trials on two measures, the average reward and the average merge-time for both agents. The results are shown in Fig. \ref{fig:timing_reward}. By reward we refer to the individual rewards $R_R$ and $R_H$ as defined in Section 2 and merge-time refers to the amount of time it takes for an agent to reach their goal lane. Reward is an obvious choice for a measure of performance because the planner aims to optimize it for the AV and assumes that the human driver has a similar goal.  
We choose merge time here as a measure of task performance because the reward function was designed to encode it but also, because it might be a better measure for what the human drivers are optimizing. We believe that people do not prefer to take the risk of waiting to the end of the road to make a lane merge and will do it early if the opportunity presents itself. So, if the lane merging goes smoothly it should lead to lower average merge times. 

\textbf{Human. } In Fig. \ref{fig:timing_reward} (a) human performance peaks when the \textit{.4R} condition is used and in (b) \textit{.6R} leads to lowest merge-time. The variation of these measures with the selfishness factor supports our hypothesis \textbf{H1} that the human performance is affected by the robot's collaboration. 
To test whether the cooperative condition \textit{.6R} outperforms the selfish \textit{1R} baseline, we compared their merge-times  using an unpaired  
Student's t-test and found that the merge-time was significantly lower for the human ($p < 0.05$) in the \textit{.6R} condition.

\textbf{Robot. } A one-way ANOVA was conducted to test the variation of the AV's merge-time with different $\alpha$ and was found to be statistically significant (F$(5, 302)$)$=20.96$, $p<0.0001$. This affirms that different levels of cooperation affect the performance of the AV. From Fig. \ref{fig:timing_reward} we observe that the two cooperative settings of \textit{.4R} and \textit{.6R} perform equally-well or even better when compared to the selfish 1R baseline for both measures, supporting our hypothesis \textbf{H2}. The observation that \textit{.6R}  improves the humans' ability to reach the goal is not surprising since it is more considerate towards the human's goal. However, it is surprising that the \textit{.6R} condition performs better even for the robot than optimizing its selfish goal. We attribute this to its possession of a more accurate model of the user's behavior. During planning, the \textit{1R} assumes that the human actively plans for the robot's goal and not her own. However, this is false and probably leads to it modifying its plan often making it suboptimal.

In Fig. \ref{fig:timing_reward}, we also observe that the human is generally slower at lane merging than the robot. This is because the AV's reward favors the fastest lane merge but the subjects were not instructed to merge as quickly as possible, they were merely asked to do it before the road ends.

\begin{table}[h]
\centering
\caption{Failure Rate to Reach Goal Lane}
\label{tab:failure}
\begin{tabular}{lllllll}
\toprule
& \textbf{0R} & \textbf{.2R} & \textbf{.4R} & \textbf{.6R} & \textbf{.8R} & \textbf{1R}    \\
\midrule
\textbf{HV} & 15.2\% & 10.3\% & 7.5\% & 4.3\% & 9.5\% & 8.3\% \\
\textbf{AV} & 52.2\% & 10.3\% & 7.5\% & 2.1\% & 6.3\% & 6.7\%
\end{tabular}
\end{table}

\textbf{Failure Rate. } Merge time was only computed for successful trials, i.e. trials where the car was able to successfully merge into its goal lane before the end of the road was reached. Table \ref{tab:failure} shows the percentage to reach the goal for the human and the AV. Again, we found that the cooperative \textit{.6R} condition outperformed all others, including the \textit{1R}, which means that both the human and AV were able to reach their goal lanes more often when the robot had a fair reward function. This lends more support to the hypotheses \textbf{H1} and \textbf{H2}. Every condition barring \textit{0R} achieves failures of $< 11\%$, supporting the argument that our framework is able to perform the task. We explain the reason for this high incidence of failures for \textit{0R} in section \ref{ssec:qualitative}.

\subsection{Subjective Measures}
\label{ssec:subjective}
\begin{figure}
\centering
\includegraphics[width=0.99\linewidth]{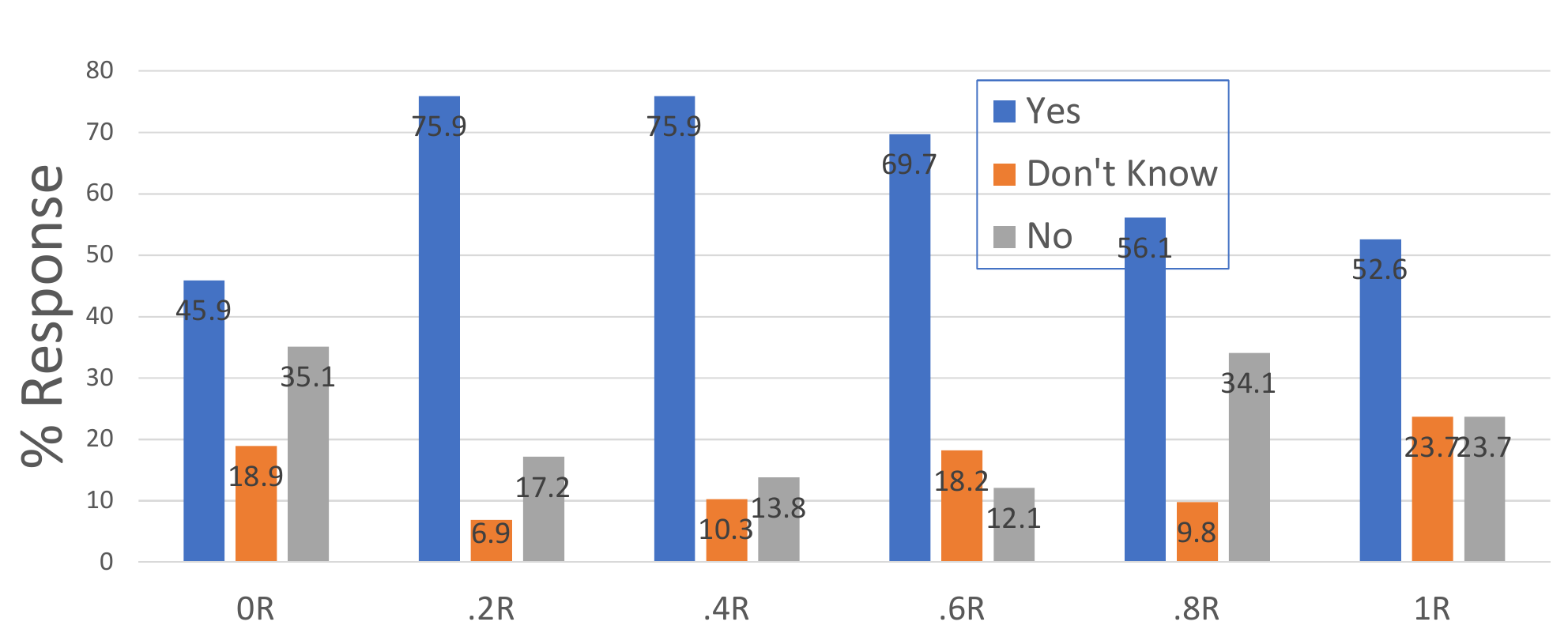}
\caption{Q1. Was it considerate of your goals?}
\label{fig:q1}
\vspace{-0.1cm}
\end{figure}

\begin{figure}
\centering
\includegraphics[width=0.99\linewidth]{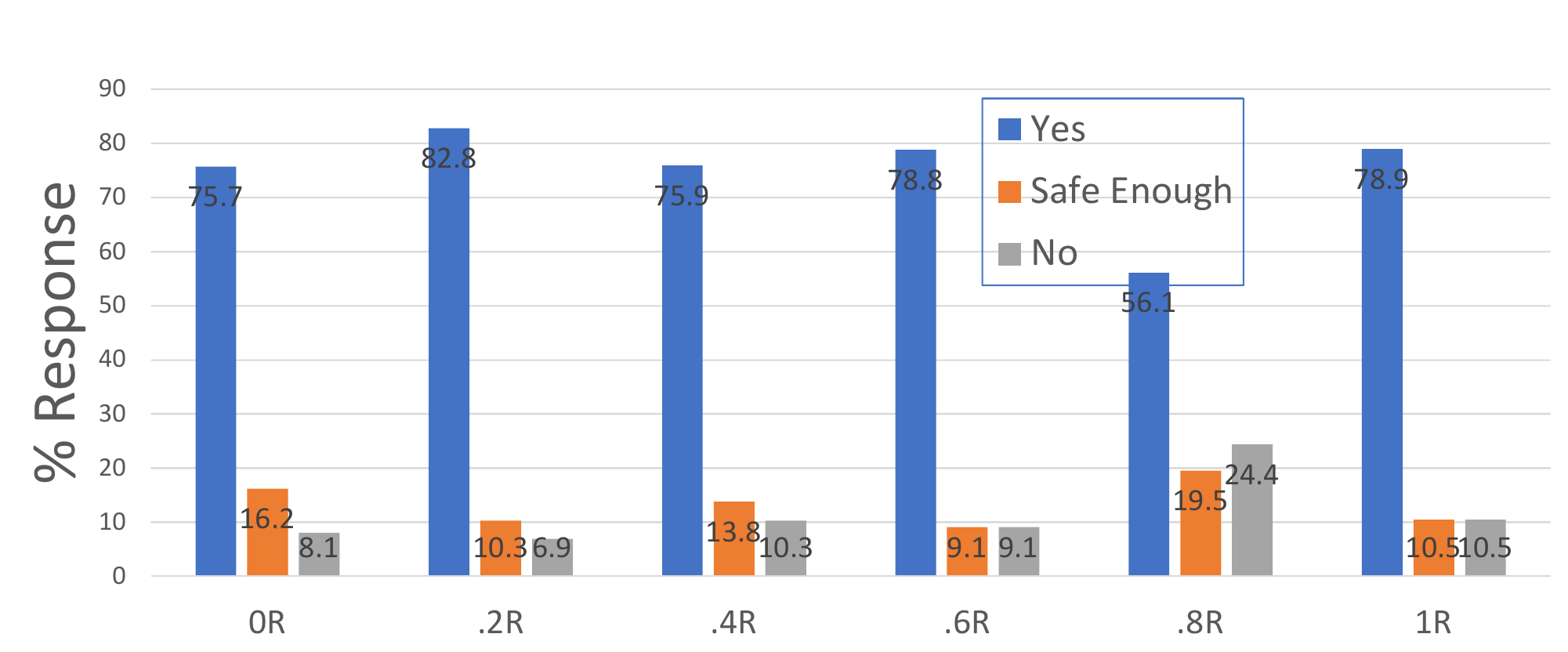}
\caption{Q2. Did the other car drive safely?}
\label{fig:q2}
\vspace{-0.25cm}
\end{figure}

After every trial, the participants were asked to answer the following two questions on a three-point scale.
\begin{enumerate}[label=\textbf{Q\arabic*. }]
	\item Was the other car considerate of your goals?
	\item Did the other car drive safely?
\end{enumerate}

For \textbf{Q1}, we explained that a considerate driver is one that changes their behavior for the benefit of other drivers. For instance, in the case of a car waiting to turn left at an intersection (with no traffic light on a U.S. road), being considerate at an intersection could mean reducing speed to give them time to make the turn or  it might be better to speed up and pass them quickly. This notion of consideration is strongly influenced by the context, e.g. depending on  relative speed, traffic condition, \etc, and can not be easily determined by a static rule. 

Fig. \ref{fig:q1} shows the response. Here, the three-point scale goes from "No", to "Don't Know", to "Yes". Excluding \textit{0R}, the percentage of considerate trials decreases with increasing selfishness ($\alpha$). Supporting the argument that less selfish agents are perceived more favorably in terms of consideration. A surprising observation is that the agent considered the least considerate is \textit{0R}, the AV which solely optimizes the human's reward. We hypothesize that this agent is not perceived as rational and we believe that people reserve the considerateness attribute for rationally acting agents. The \textit{0R} agent might be considered irrational because it does not act in its own self-interest when given the chance, for instance, even after the human has made a lane change and the agent is free to merge into its target lane it does not do so and seemingly takes random actions. This is illustrated in Fig. \ref{fig:alpha_0} and the next subsection gives more details on why this occurs. We believe that rationality is a prerequisite for being perceived considerate. 

\textbf{Q2} gauges whether the subject felt safe during the interaction on a three-point scale, from "No", to "Safe Enough", to a "Yes". We wanted the participants to use the notion of safety that they have developed by real-world driving and so did not provide examples of safety for the scenario. The results are grouped by the selfishness factor ($\alpha$) in Fig. \ref{fig:q2}. Apart from \emph{.8R}, every condition seems safe and there is no clear trend. This leads us to believe that safety is difficult to judge in this environment due to the detachment caused by the top-down view as opposed to a first-person view and in the future, a more involved simulation set-up might be able to recover it.

\subsection{Qualitative Analysis}
\label{ssec:qualitative}
\begin{figure}
\begin{tabular}{cc}
\includegraphics[width=0.45\linewidth]{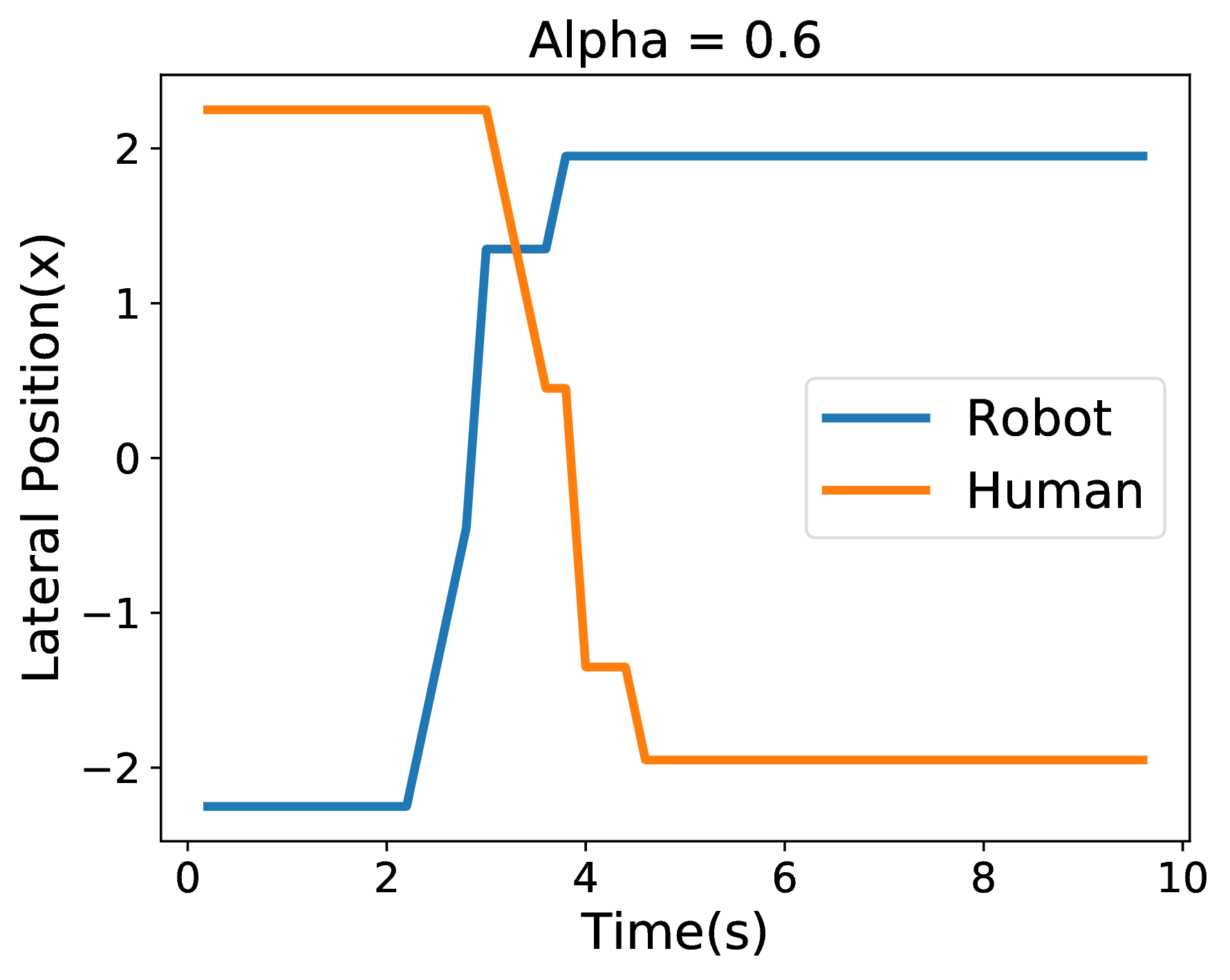} &
\includegraphics[width=0.45\linewidth]{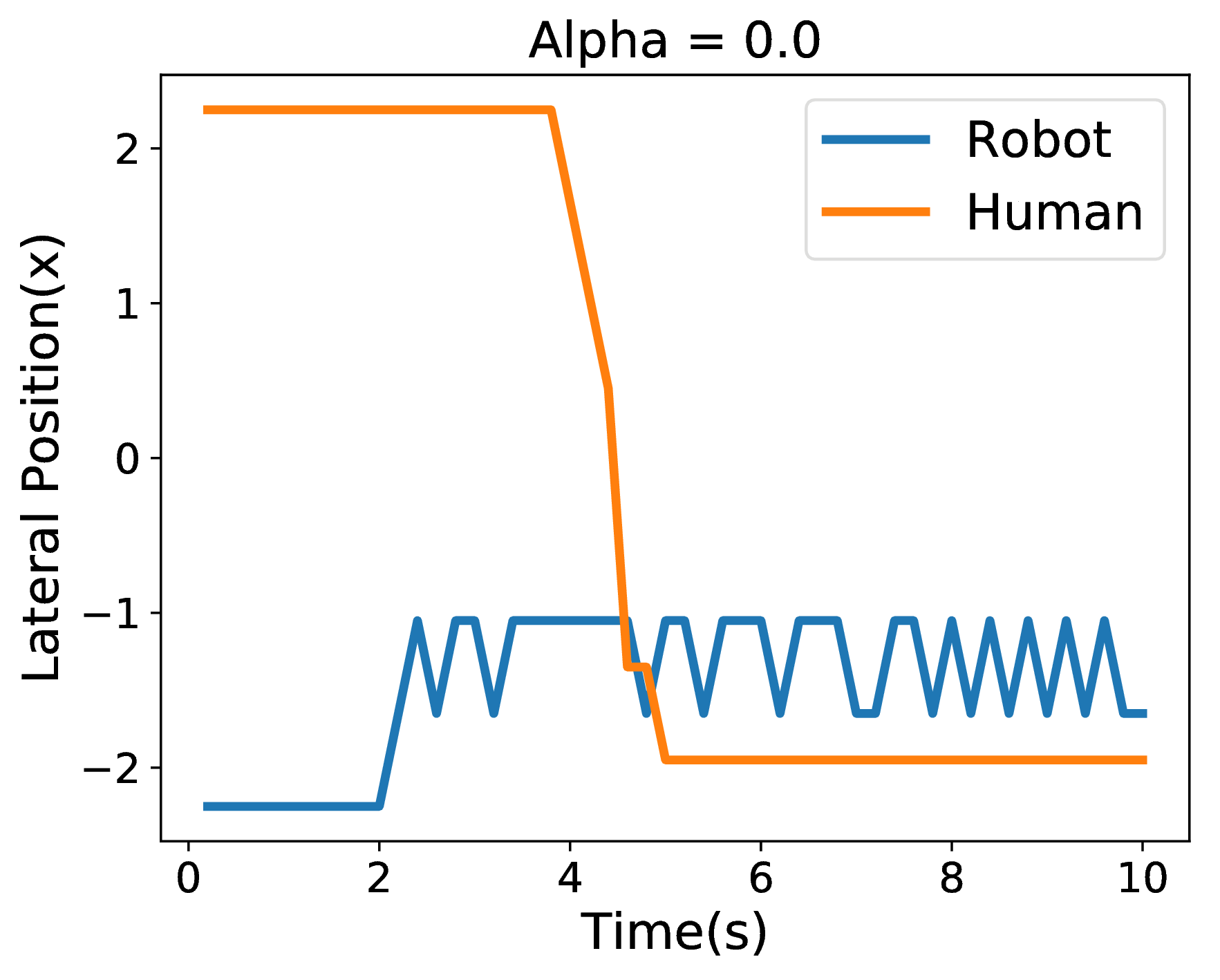} \\
(a)  & (b) 
\vspace{-0.2cm}
\end{tabular}
\caption{Effect of $\alpha$. The lateral position is shown for agents \textit{.6R} and \textit{0R} for two different trajectories in (a) and (b) respectively. (a) is a typical lane change sequence where both agents successfully steer to their goal lane. While in (b), the human again completes their merge successfully, but the AV takes random steering decisions making it appear irrational.
}
\label{fig:alpha_0}
\end{figure}

\begin{figure}
\begin{tabular}{cc}
\includegraphics[width=0.45\linewidth]{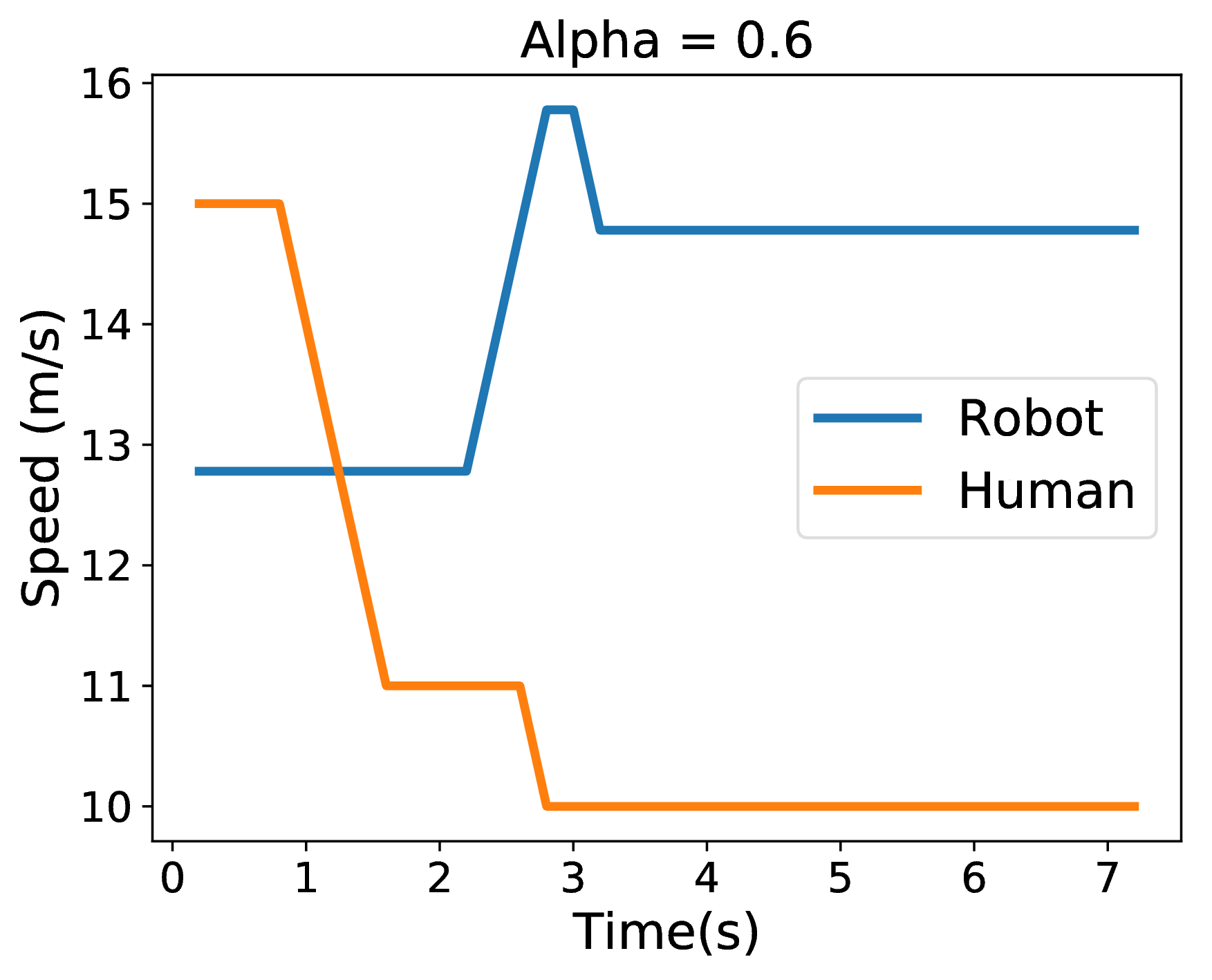} &   \includegraphics[width=0.45\linewidth]{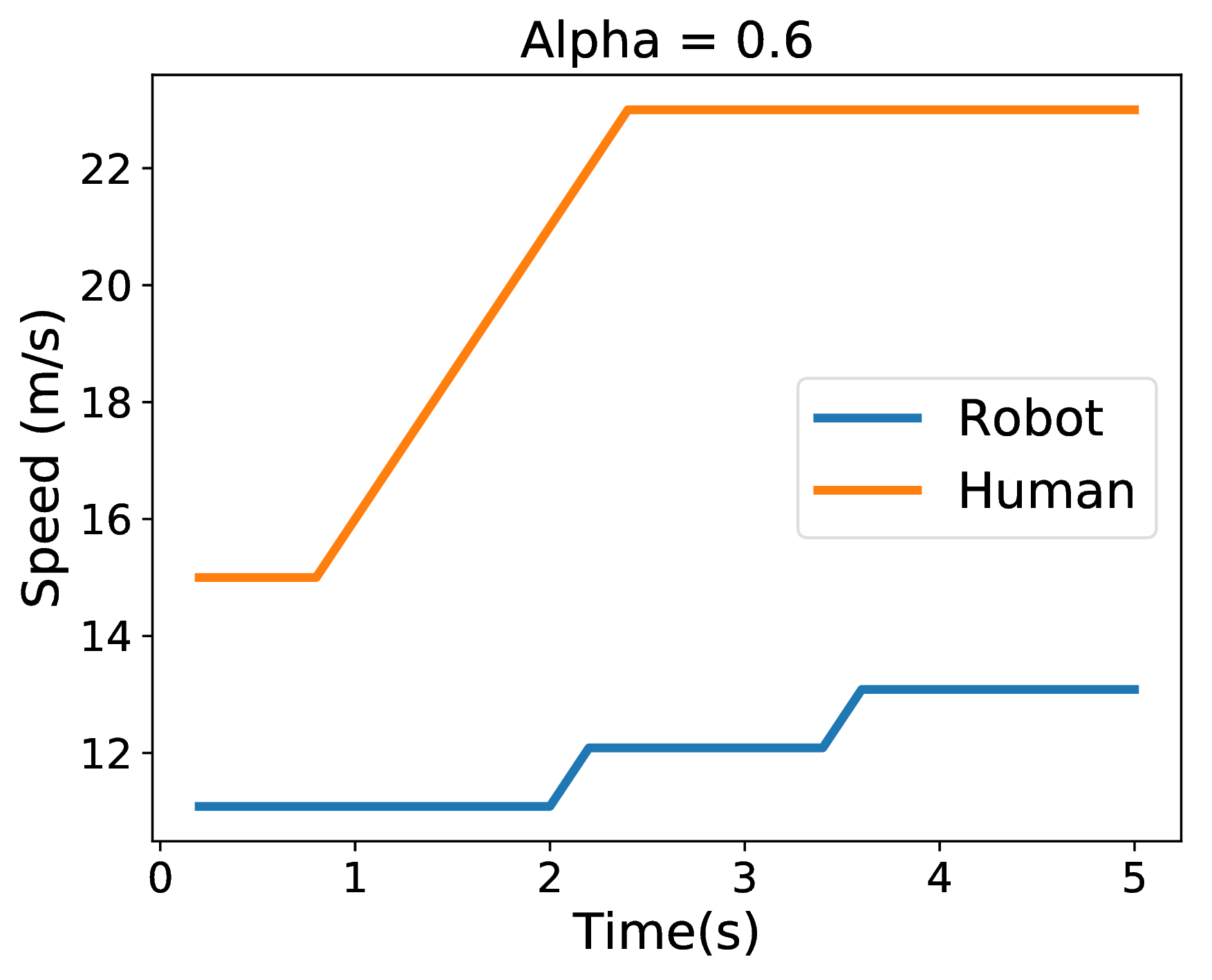} \\
\includegraphics[width=0.45\linewidth]{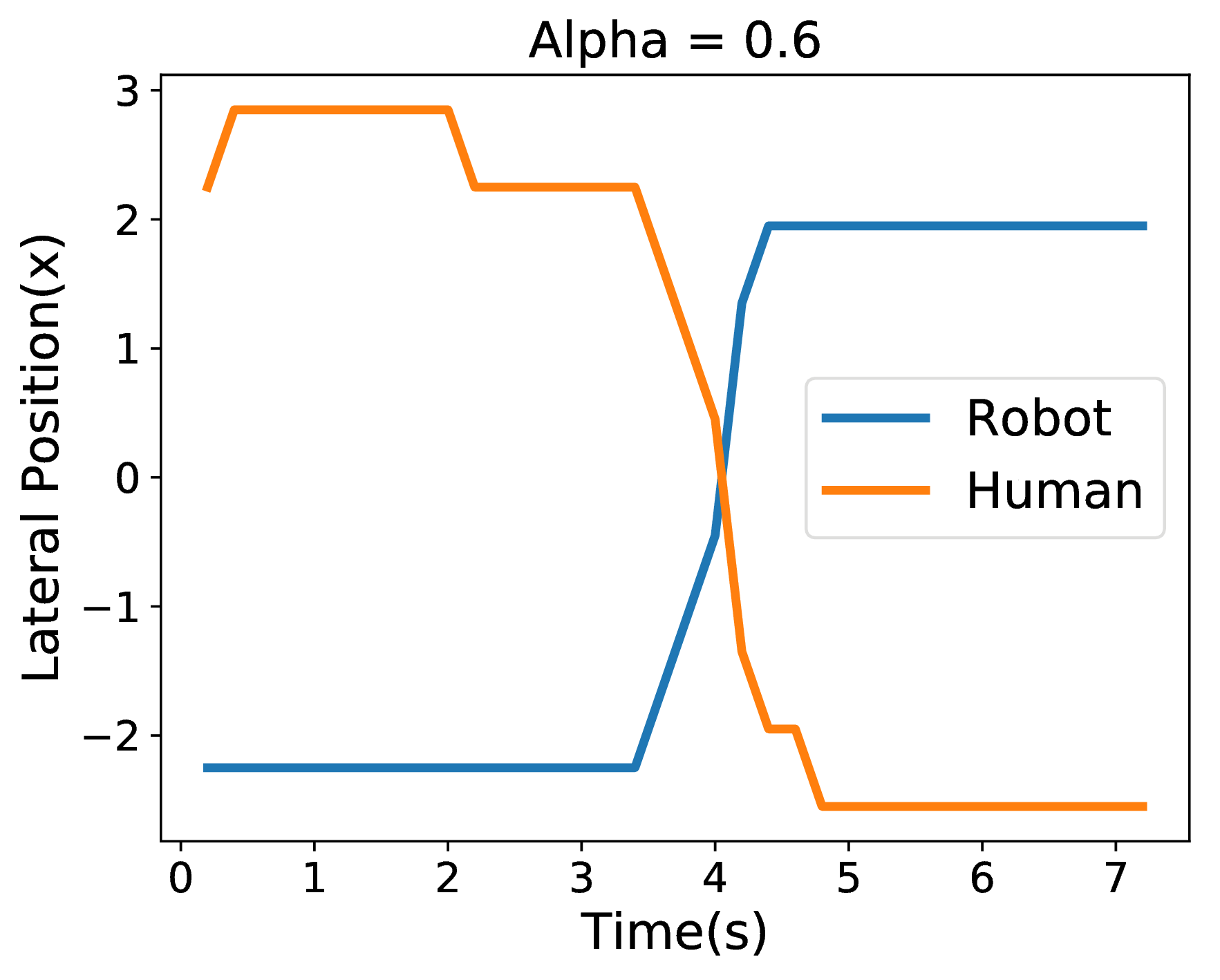} &   \includegraphics[width=0.45\linewidth]{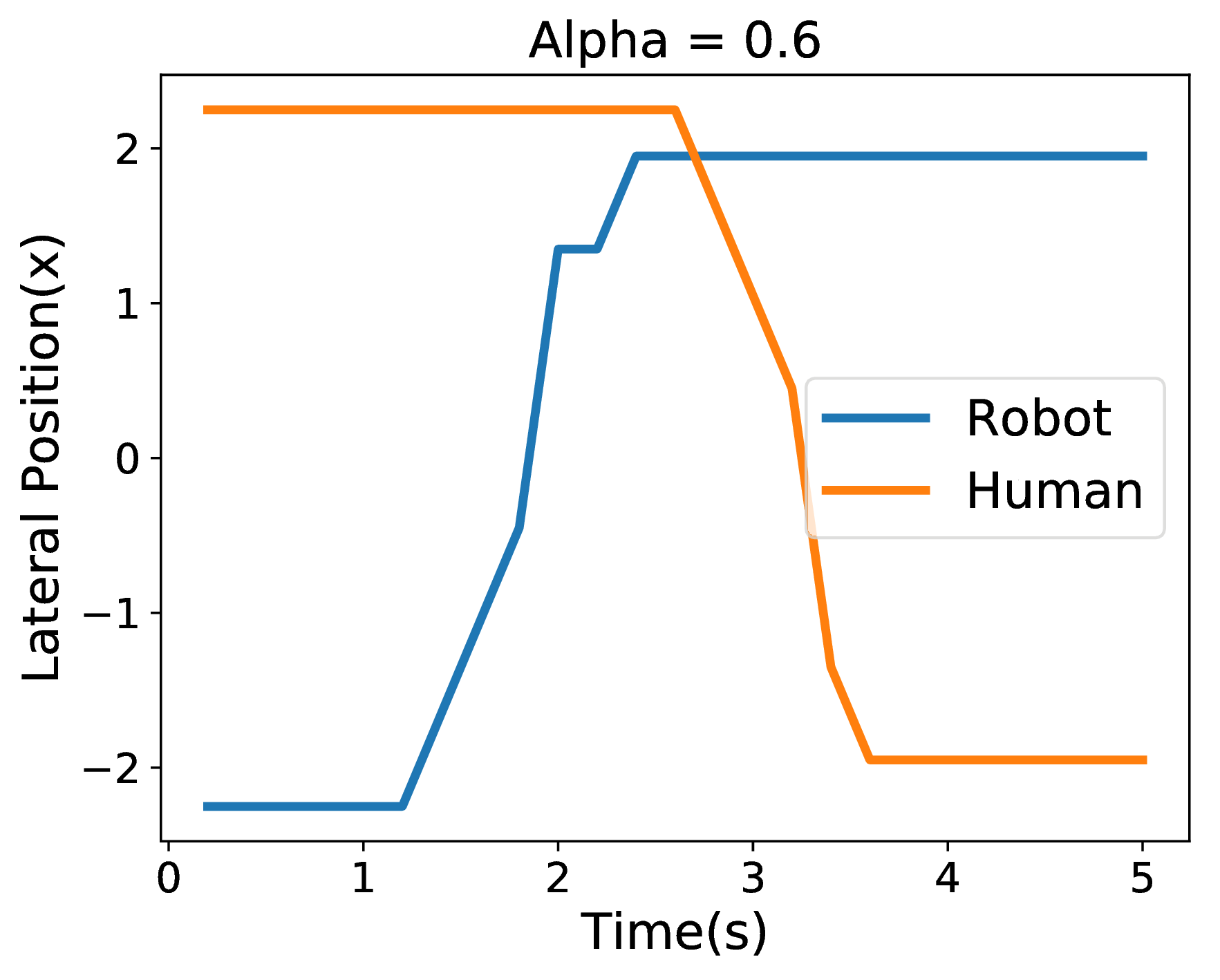} \\
(a)  & (b) 
\vspace{-0.2cm}
\end{tabular}
\caption{Adaptation to human actions. The first and second rows plot longitudinal speed and lateral positions respectively. (a) and (b) are two trajectories. In (a), the human brakes and slows down so the AV accelerates to get ahead performing the merge afterwards. In (b), the human accelerates getting ahead of the AV allowing it to start steering early because it does not have to use the accelerate action first. Since the initial speeds and selfishness factor were the similar across columns, the difference in the AV's behavior can be attributed to the human's actions.
}
\vspace{-0.2cm}
\label{fig:adapt_human}
\end{figure}

In this section we illustrate the behavior of our method through select examples. In particular, we explore the effects of the selfishness factor $\alpha$ and of the human driver's behavior in Figs. \ref{fig:alpha_0} and \ref{fig:adapt_human} respectively.   
Fig. \ref{fig:alpha_0} plots the lateral position ($x$) of both vehicles with respect to time for two trajectories. In (a), a typical scenario of a successful merge is shown, where both the human and AV start steering towards their goals and after a short time period, reach them. In (b), the unselfish \textit{0R} steers towards its goal lane initially but then starts steering back and appears to be taking random actions. This is because \textit{0R} optimizes only for $R_H$, i.e. for this AV's reward, while the human driver reaching their goal lane is ideal, their own lane does not matter. This leads the planner to oscillate between choosing equally rewarding plans that may lead to either lane. The human in this scenario is aware of the AV's goal lane, but not of its peculiar reward function, and so might conclude from this behavior that the agent is irrational. We used this argument in Section \ref{ssec:subjective} to explain why participants in our study did not find the \textit{0R} agent to be considerate. This behavior explains the high high failure rate for the AV, and it's departure from the human's expectation might also explain the higher failure of the HV, for \textit{0R} in Table \ref{tab:failure}.

Fig. \ref{fig:adapt_human} highlights the capability of our approach to adapt to the human driver's behavior. Columns (a) and (b) are two trajectories that were performed with the same \textit{.6R} selfishness factor and similar initial conditions but show different behavior due to the difference in human actions. 

\vspace{-0.1cm}
\section{CONCLUSIONS AND FUTURE WORK}
\label{sec:conclusion}
We proposed a collaborative method for mixed-autonomy driving that optimizes a collective reward and showed that it can adapt to the human's behavior and find feasible solutions. To do this, we were able to use an approximate reward model for the humans due to their adaptability and did not require a large corpus of demonstrations to learn from or hand-code specific behaviors. Manipulating a single parameter in our framework gave rise to different levels of cooperation and we found that optimizing for a fair-share of the reward led to decreased merge times and failure rates. Providing evidence for the utility of collaboration in this domain and affirming our assumption that driving was a non-zero-sum game. 

\textbf{Future.} We found that subjective notions of human safety are difficult to judge in this setup and hope to see future work utilizing Virtual Reality to create realistic environments to measure it. 
An interesting avenue would be to personalize the selfishness factor and estimate it in an on-line manner for every interaction.
The idea of collaborative planning can be applied to other domains where agents interact to achieve personalized goals, for instance, factory robots sharing tools with a human worker. 
We think the idea of controlling the meta-behavior of an AV through variables like a selfishness factor can be used to personalize an autonomous car to the comfort of its rider. For example, some people might prefer an AV that yields easily to other drivers and is generally more considerate, but the same person may choose to change this behavior in situations where they are pressed for time. 

\addtolength{\textheight}{-12cm}   






\section*{ACKNOWLEDGMENT}
We would like to thank colleagues at HRI - Priyam, Mustafa, Trevor, Reza and Sasha for insightful discussions. 
We would also like to acknowledge Amirreza, Ching-An, Brian, Byron, Himanshu, Kalesha, Sonia, Sid, Sangeeta, and Shubh for their valuable suggestions and comments.


\vspace{-0.3cm}

\bibliographystyle{IEEEtran}
\bibliography{refs}  


\end{document}